# Lessons learned developing and using a machine learning model to automatically transcribe 2.3 million handwritten occupation codes

Bjørn-Richard Pedersen[1], Einar Holsbø[2], Trygve Andersen[1], Nikita Shvetsov[2], Johan Ravn[3], Hilde Leikny Sommerseth[1], Lars Ailo Bongo[2*]

[1] Norwegian Historical Data Centre, UiT The Arctic University of Norway
[2] Department of Computer Science, UiT The Arctic University of Norway
[3] Medsensio AS, Tromsø, Norway
[*] Corresponding author: lars.ailo.bongo@uit.no

**Abstract**

*Machine learning approaches achieve high accuracy for text recognition and are therefore increasingly used for the transcription of handwritten historical sources. However, using machine learning in production requires a streamlined end-to-end pipeline that scales to the dataset size and a model that achieves high accuracy with few manual transcriptions. The correctness of the model results must also be verified. This paper describes our lessons learned developing, tuning and using the* Occode *end-to-end machine learning pipeline for transcribing 2.3 million handwritten occupation codes from the Norwegian 1950 population census. We achieve an accuracy of 97% for the automatically transcribed codes, and we send 3% of the codes for manual verification. We verify that the occupation code distribution found in our results matches the distribution found in our training data, which should be representative for the census as a whole. We believe our approach and lessons learned may be useful for other transcription projects that plan to use machine learning in production. The source code is available at: https://github.com/uit-hdl/rhd-codes*

# Introduction

Over the last few decades, we have witnessed a boom in the digitization of historical documents, and many national archives are developing services where the public can easily access their cultural heritage. For example, in the Norwegian state budget for 2019, 140 million NOK were allocated to the National Archive of Norway. About half of the budget was assigned for further development of the Digital Archive, which stores and distributes digitized historical documents. This will increase the number and availability of digitized historical documents. However, there is an increasing demand from the research community to have these documents in a data format suitable for research based on data analysis. This demand has yielded the development of more time and cost efficient systems in recent decades. Of particular interest for our project is the development of automatic text recognition for population data, typically characterized as handwritten documents with a tabular structure. These sources form the basis in the construction of the Norwegian Historical Population Register (HPR; http://www.rhd.uit.no/nhdc/hpr.html).

The HPR will include the records of the 9.7 million people who lived in Norway in the period from 1801 to 1964. We are building life trajectories across multiple generations by linking individual and contextual attributes derived from population censuses and church records. In the HPR, we have over 10 million manually transcribed person entities from the 19th and early 20th century population censuses, and approximately 20 million person entities from church books. While the censuses have primarily been transcribed by professional transcribers, the church books have been transcribed by a mix of volunteers and professionals. Selected columns in the church books have recently been outsourced through a manual transcription agreement between the National Archives of Norway and three commercial genealogy companies, and 59 million person entities were transferred to HPR in October 2021. However, we still have an enormous amount of data awaiting transcription and integration into HPR, so automation would undoubtedly be both time and cost efficient.

Handwritten text recognition (HTR) methods have gained increased attention over the past two decades, and different approaches have been proposed for both online and offline automatic and semi-automatic transcription of handwritten text, character and/or digit recognition (Bottou et al., 1994; Ghosh & Maghari, 2017; Liu, Nakashima, Sako, & Fujisaw, 2003; Plamondon & Srihari 2000). We describe an offline solution to automatically transcribe handwritten single and multi-digit numbers from the Norwegian full count 1950 population census. This was the last census where the information was aggregated manually, and it is therefore an important bridge to the later electronic censuses. The census manuscripts have been scanned by the National Archives of Norway. The questionnaires include two stippled rows for each individual registered (Figure 1). In the first row, the enumerator registered the characteristics of each person, divided into 26 columns, while the second row was kept blank. This row was strictly reserved for the statistical preparation of the census, and Statistics Norway has greyed out the row and typed a warning message: "Skriv ikke på denne linjen" (Do not write on this line). The second row consists of numbers written with a coloured pencil (usually red), by employees at Statistics Norway or their liaisons in the different Norwegian counties. These numbers represent codes for family position, de facto/de jure residency, marital status, occupation and education.

Handwritten digit recognition is a textbook example in machine learning, with numerous tutorials available for different machine learning frameworks. The MNIST dataset of 70,000 handwritten single digits (Lecun, Bottou, Bengio, & Haffner, 1998; http://yann.lecun.com/exdb/mnist/) was an early benchmark dataset in machine learning, which was widely used in many research papers ten years ago, and the best algorithms achieve almost perfect accuracy today (e.g. the 0.23% error rate found in Cireşan, Meier, & Schmidhuber, 2012). There are numerous solutions for automatic detection of numbers in images, such as reading street numbers. However, existing solutions trained on MNIST and other modern datasets may not work well for historical data, because the digits in these are written in many other ways (Kusetogullari, Yavariabdi, Cheddad, Grahn, & Hall, 2019).

*Figure 1*: *A zoomed-in view on the Occupational columns in the census (Supplementary Figure 1 shows the entire census page). The column of interest in this paper has been highlighted in yellow. The header text is: Self-employed, clerks, workers and servants state their position or their type of work. Unemployed, active-duty military and other temporarily unemployed state their normal position. Others state their primary means of making ends meet or what they do.[1]*

To implement a transcription solution, we must first select software libraries, services or tools that achieve high accuracy for our census data. Digit recognition can be implemented and trained from scratch using a deep learning framework such as Keras (https://keras.io/) or PyTorch (https://pytorch.org/). We can tailor the digit recognition for our data by implementing a model that takes into account that our codes always have a specified number of digits and that they are written in most cases using a red pencil. However, implementing a model from scratch requires machine learning expertise, as well as a labelled training dataset. It is therefore easier to use a cloud service for handwriting recognition such as the Microsoft Azure Computer Vision API (https://azure.microsoft.com/en-us/services/cognitive-services/computer-vision/) or Google Cloud Vision (https://cloud.google.com/vision/). Their models are pre-trained using a very large dataset, but it is possible that accuracy with respect to a few digits will suffer due to the model being general for all handwritten text. There are standalone tools that convert handwriting to text such as Evernote (https://evernote.com/) and Microsoft OneNote (https://www.onenote.com/), but these are also generalized to recognize all text, and because they are designed to work on one page at a time, it would be challenging to customize these services for the recognition of millions of images. Finally, there are transcription platforms developed for historical documents such as Transkribus (Muehlberger et al., 2019), which have been shown to work well for historical documents and that allows training and using a data specific model.

---

[1] Full translation: Self-employed, clerks, workers and servants state their position or type of work, for example: tenant, farming at home, wholesaling, iron turner, private driver, etc. Unemployed, active duty military and others temporarily unemployed state their normal position. Others state what they do for a living or what they occupy their time with (old age pension, *'føderåd'* [retired people with special benefits from the farm], personal wealth, pension, etc., or housewife, housework at home, student etc.).

In this paper, we describe our *Occode* machine learning pipeline for tabular data transcription and our experiences using it in a production setting. First, we implemented an end-to-end pipeline for image cutting, image cleaning, transcription and verification. This pipeline ensures reproducible and scalable processing of image data. Second, we tune the model to achieve the high accuracy and few manual transcriptions required to transcribe the 2.3 million entries for 3-digit occupation codes in the Norwegian 1950 census. We achieve an accuracy of 97% for the automatically transcribed codes, with 3% sent for manual validation and correction. Third, we propose methods for verifying the correctness of the transcribed results. We did not find systematic errors in the transcribed results, which will be implemented in the HPR.

# Methods

## Occode: An end-to-end machine learning pipeline for transcription of tabular datasets

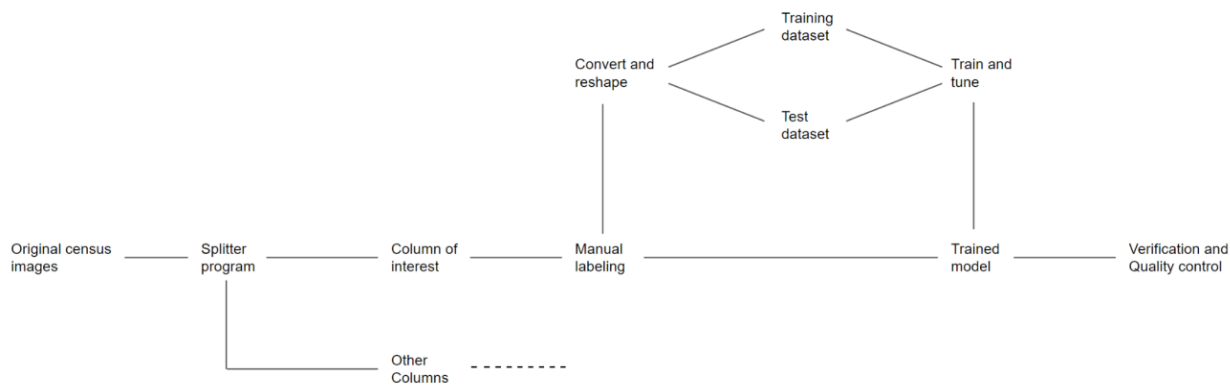

*Figure 2*: *The Occode end-to-end machine learning pipeline for transcription.*

We developed an end-to-end machine learning pipeline for transcription of tabular data (Figure 2). First a splitter program divides the scanned questionnaire tables into individual cells (Supplementary Figure 1). The cell contents are stored as separate images. The images from one column comprise a dataset. A dataset typically contains millions of small images. To avoid cutting text written outside the cell boundary box, the splitter program extends the cell boundary if it detects pixels of certain colours on one of the cell borders.

We implement a workflow for labelling a randomly selected subset of the images in a dataset. The human transcriber is shown images in a Graphical User Interface (GUI) and is asked to label them. The GUI may simply show an image and record the user's input, but it may also use a clustering or classification method and ask the user to select the images not belonging to the class. The result is a labelled dataset in which each image is labelled by one or more persons.

We develop or adapt a machine learning model for each column that we want to transcribe. The model may be provided by a cloud service, a machine learning framework or library or an open source code repository. We may also develop a new model from scratch. To achieve the best results, we use a labelled dataset split into a training set and a test set. We ensure that the model is trained with the same distribution of values as in the full dataset. We may be able to reuse a labelled dataset to train a model for another column, for example if both columns are numerical, but it is often necessary to adjust the images used for the training to ensure that they reflect the value distribution of the new column.

The images are preprocessed to make them as uniform as possible before they are sent to the model for training. First, the images are converted to greyscale and assigned the correct data type, then they are resized to a standard height and width. After these steps, the model is trained, and its performance can be evaluated on the testing set. Finally, we use the trained model to transcribe the full dataset for the column, and then verified the quality of the transcribed data.

## Occupation codes in the Norwegian 1950 population census

The 1950 census comprises 801,000 double-sided questionnaires, resulting in about 1.6 million scanned images. The image sizes are large rectangles in format (29.7×70.7cm), with high resolution (8,500×3,600 pixels). We used our splitter program (described above) to generate a dataset for the occupation codes column. We successfully split 92.83% (714,904 out of 770,094) of the questionnaires. The result was 7,342,113 images, of which 5,021,637 were empty rows and 2,320,476 images had codes or invalid text. We called the 2.3 million images the *full* dataset.

## Labelled training datasets

We created a 3-digit occupation code training dataset based on a 2% random sample extracted by Statistics Norway, by selecting every 50th page after the completion of the census enumeration. Of the split images, 36,065 images were in the 2% sample and therefore used in our *2% training* dataset. These were manually transcribed by the Norwegian Historical Data Centre. Each image was labelled by a professional transcriber.

We also created a *3×1-digit training* dataset in which we split the 3-digit code numbers into single digits. We used a simple approach for segmentation by fitting a three-component mixture of normal distributions to a normalized histogram describing the marginal ink density on the x-axis. If an image did not contain three modes, the image was skipped, because it did not identifiably contain the expected number of digits. We discarded 795 3-digit images that our method did not split into 1-digit images. The resulting *1-digit training* set consisted of 102,809 images. We labelled these using the 3-digit labels.

Finally, we created a third *random sample training* set by randomly selecting 30,000 images from the 2.3 million images in the full dataset. We used our predictions from a model trained on the 2% training set to predict labels that were then manually validated or corrected by one of ten transcribers.

## Verification dataset

We created and labelled a 3,000 image *verification* dataset by randomly selecting 3,000 images that were not in any of the training datasets. We manually verified that we had at least 80 unique classes in the set, and that all images contained a 3-digit code.

## Machine learning models for multi-digit classification

We evaluated several models for our occupation code transcription using the *2% training* set. First, a pre-trained model using the Microsoft Azure Compute Vision application programming interface (API). Although the service is easy to use, the accuracy of the model was not high enough for our 3-digit codes. Second, the Azure Custom Vision API (version 1.1 preview) allowed us to create classifiers from scratch; however, limits on the number of images that could be used to re-train the model reduced its performance, and a limitation of 50 different classes in Custom Vision API made it impractical for our purpose. We therefore decided to develop and train our own model from scratch using our own training data.

We developed three models in Keras (https://keras.io/) and trained them using labelled 3-digit occupation code images. The first model *(3×1-digit model)* split the code into three separate digits and then classified each separately using a single-digit classifier trained on single digit images. The second and third models *(3-digit* and *3-digit-sequential model)* were trained on the labelled 3-digit occupation codes. We tuned the models by testing different hyper-parameters.

In summary, we used both a Convolutional Neural Network (CNN) and a Recurrent Neural Network (RNN) that views an image as a matrix of numbers and are trained to extract the most important features, such as shapes, in the matrix to classify the objects in the image into a class. We refer to Hirschberg and Manning (2015) for an overview of the use of such deep learning methods for natural language processing.

The *3×1-digit model* consists of 4 convolution layers with a pooling layer in between each convolution layer, filter sizes starting at 32 and increasing to 128 for the last two layers, kernel size of 3×3, with a Rectified Linear Unit (ReLU) activation function. Two fully connected layers with filters 512 and 10 calculate the final classification into the correct class 0–9. The model utilizes Sparse categorical cross entropy and the Adam optimizer with a learning rate of 1e-04.

The *3-digit model* is similar to the *3×1-digit model*, but it has an additional convolution and pooling layer with a filter size of 256. There is also another fully connected layer with filter size 256, and the output is from 0–264 instead of 0–9. This model uses sparse categorical cross entropy and the Adam optimizer and has a learning rate of 0.001.

The architecture of the *3-digit-sequential model* consists of a combination CNN and RNN. It has 2 convolutional layers, filter sizes of 32 and 64, with pooling layers in between. Then a fully connected layer with filter size 64, and a 20% dropout layer. Two LSTM layers make up the RNN part of the network, before a final fully connected output layer that classifies the input into 1 of the 13 possible characters that can make up our 3-digit sequence. The model uses the Connectionist Temporal Classification (CTC) loss function (Graves, Fernández, Gomez, & Schmidhuber, 2006), and performs Sequence-to-Sequence classification instead of the One-to-One classification used in the two previous models. The model therefore splits an image into time-steps and performs classification for each time-step. In addition to the 10 digits, there is also a class 'B' for empty images, 'T' for images containing text and the pseudo-blank '-', spacing character that is a built-in feature of the CTC loss function. This model also uses the Adam optimizer. We used Early Stopping with a patience value of 10 to prevent overfitting the model to the specific images in the training set. So, if the model does not see improvement in a span of 10 training iterations, it ends the training and retains the training values from the peak 10 iterations previously.

# Results

We have three main requirements for the machine learning model:

1. The accuracy of the automatic transcription should be similar to human level performance: 97%.
2. The number of images sent to manual validation and correction should be less than 100,000 (4.31% of the valid code images).
3. In the resulting automatically and manually transcribed occupation codes, the distribution of the codes for each occupation should match the distribution found in our training data, which we assume is representative for the census as a whole

The first and second goals are in conflict, because a higher accuracy requires more manual transcription. We can tune this trade-off by adjusting the confidence required for a code to be automatically transcribed. A higher confidence threshold improves accuracy but requires more manual validation and correction. The third requirement is to ensure that the model does not systematically introduce errors into the transcribed data.

Although we only used the *3-digit sequential* model in production, we include the negative results for the other two models because we believe these provide useful lessons learned in how to discover, analyse and solve common challenges when using a machine learning model to transcribe a large dataset.

## Decision analysis to choose model

Before applying the model to transcribe the occupation codes, we needed to choose between the *3×1-digit* model, *3-digit* model or the *3-digit-sequential* model. We started by comparing the *3×1-digit* and *3-digit* models, because they use a similar deep learning architecture. For both, we used k-fold cross validation with an 80–20 split of the *2% training* dataset into training and test.

The *3-digit* and *3×1-digit* models output vectors of confidence scores for each class. We used these scores to decide which class the image should be classified as. For our purpose, the decision took the form "assign to a given image the class $k$ if the confidence in $k$ is above the threshold $t$". If there is no $k$ for which the confidence is higher than $t$, the model cannot discriminate between classes, and the image must be sent to manual transcription. For the *3×1-digit*, model this decision was made three times, once for each digit. If it is the case that any one of these numbers could not be classified, the whole code could not be classified and therefore must be sent to manual transcription. Hence the choice of a $t$ is a trade-off between error and manual transcription cost: a higher $t$ leads to fewer model-induced errors but also leads to more images being sent to manual transcription.

Although the confidence output by our models takes the form of a probability, we did not expect this number to be either calibrated or interesting in itself. What is interesting, however, is how many images needs to be sent to manual validation and correction to achieve a certain level (lack) of error. We expect a human transcriber to make an error in no more than 3% of the images. Because we expected to transcribe most images automatically, the relative error from manual validation and correction should be negligible. Figure 3 shows the errors from the *3-digit* and *3×1-digit* models as a function of how many images had to be sent to manual validation and correction. The uncertainty intervals are derived by treating the number of errors for each threshold as coming from a binomial probability model.

It is clear that the *3-digit* model is better than the *3×1-digit* model for almost any decision we could make. Part of this is because a certain proportion of images do not split cleanly into three single-digit numbers and hence must be sent directly to manual validation and correction (or otherwise be dealt with outside the *3×1-digit* model). However, if these images could be split and not counted against the proportion sent to manual validation and correction, the 3-digit model would still be better for almost any decision.

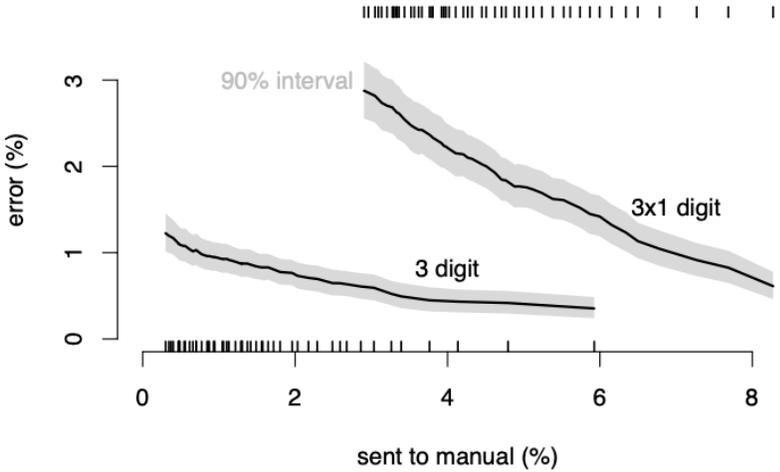

*Figure 3: 3-digit model is both cheaper and more accurate for the 2% training set.*

## Model performance does not transfer to the full dataset

Our results show that the *3-digit* model achieves the required accuracy while keeping the number of images sent to manual labelling low (Figure 3). We therefore used the model to transcribe all 3-digit occupation codes (except those in the *2% training* set). To validate the results, we plotted the distribution of classes in the training set against the distribution of classes in the classification results. Our results show a significant difference in the distributions (Figure 4), because the model overestimates the frequency of some of the larger classes (they appear above the diagonal line), and underestimates the frequency of over half of the smaller classes (they are under the line). The model performance therefore does not transfer to the full dataset. This could be due to the model being overfitted to the training data, or the training data not being representative of the full dataset.

We could not find any errors in our code, or data leaks, so we decided to test the representativeness of the *2% training* dataset. We therefore randomly selected the same number of code images as in the *2% training* dataset (36,065) from the full dataset. We used both datasets to train a binary classification model using 80% of the images in each set labelled either as *2%* or *full.* We then used the remaining 20% to test if the model could distinguish between the two datasets. Our results show that the model achieves an F1 score of 0.86 for the 2% labelled images, and an F1 score of 0.88 for the full labelled images. These results indicate that the predictions done by the binary model were not random, and therefore that the 2% images differ from the full dataset images even if the differences are not visible to the human eye. We therefore labelled a new training dataset using the randomly selected images (the *random sample training* set described above). However, the *3-digit* model trained on the *random sample training* dataset did not achieve the same performance as when trained on the *2% training* set (precision = 0.54 and recall = 0.52 vs. precision = 0.91 and recall = 0.92 for respectively the *2% training* and *random sample training* dataset). We therefore decided to implement the new *3-digit-sequential* model.

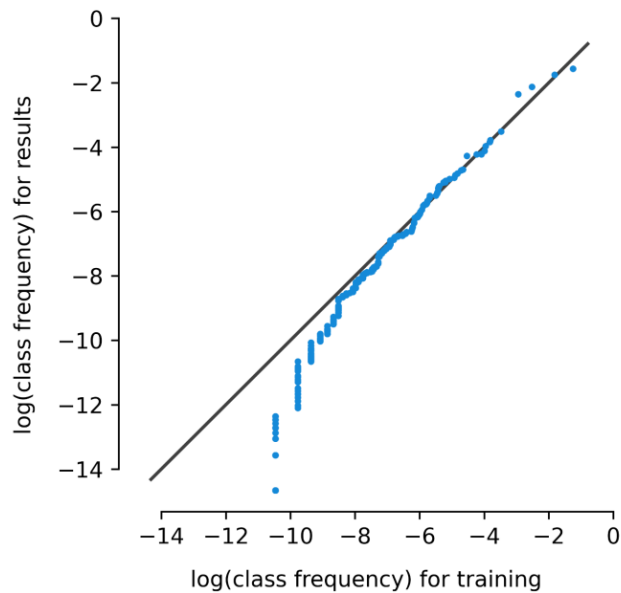

***Figure 4:*** *Confidence scores for validation set vs full dataset. The diagonal line symbolizes an even distribution.*

## New sequential model performance transfers to full dataset

The *3-digit-sequential* model (described above) achieves much better performance on the verification and test sets with a precision of 0.95 and recall of 0.95 for the validation set, and 0.96 precision and 0.96 recall for the testing set (90–10 split into train and test, no cross validation). The model performance also transfers to the full dataset (Figure 5). The distribution of classes is similar, except for the smallest classes, which only make up approximately 0.5% of the total dataset (Figure 6, left). The model performance is therefore good enough to be used for automatic transcription of the full dataset.

We believe the *3-digit-sequential* model has three main advantages over the *3-digit* model. First, the model handles blank or invalid images better. The full census consists of 7,342,113 images, of which 5,021,637 are blanks, with the remaining 2,320,476 images having codes or invalid text, so this ability to properly handle blank and text images is very useful. The *3-digit* model blank was not trained to classify blank and invalid images, because we assumed these would be accurately removed during data cleaning. However, many blank images have a diagonal line in them which prevented this. The sequential model handles such noise in the images, because they span all time steps of the image. Second, the sequential model reduces the number of classes from 328 to 13, and thereby also the class imbalance problem (Figure 6).

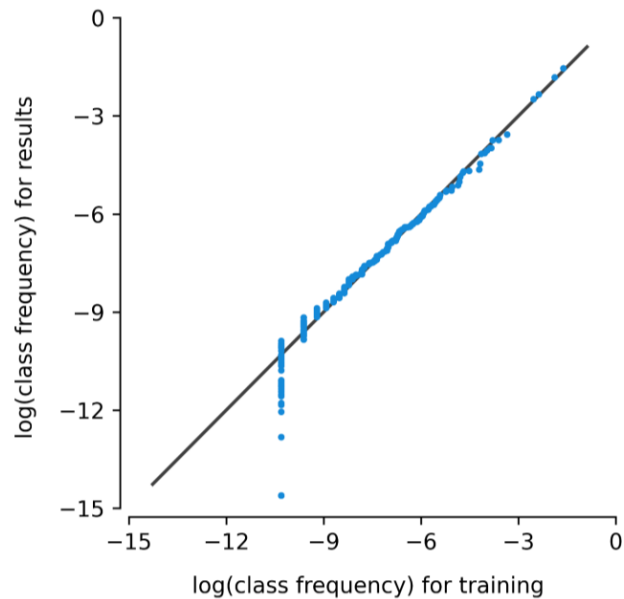

*Figure 5:* *Class distribution found in the training set is on the x-axis, and the distribution found in the predictions for the full dataset is on the y-axis. The diagonal line symbolizes an equal distribution.*

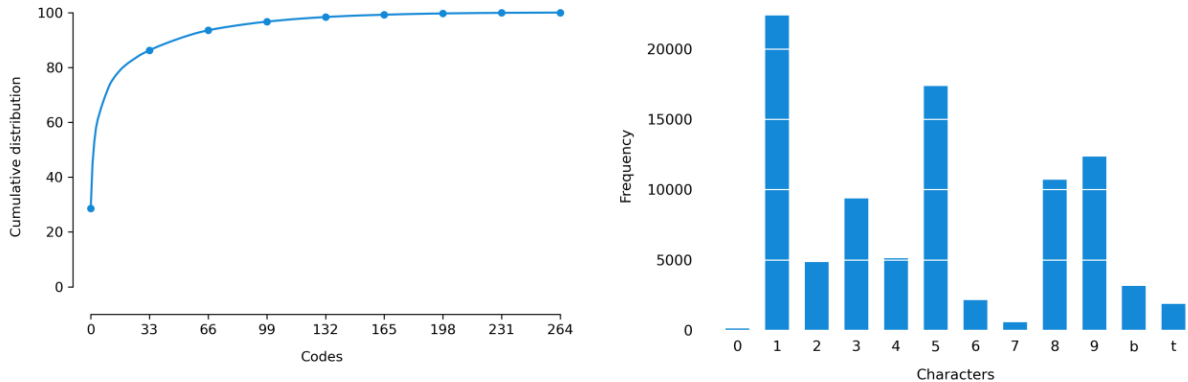

*Figure 6:* Left: Cumulative distribution of 3-digit codes in our 2% training set. The codes are sorted by most frequent first, and the x-axis shows the number of codes included in the cumulative distribution. Only 264 of 328 valid codes are in the training set. We assume the remaining codes have low frequency. Right: The distribution of single digits, blanks and text in the random sample training set.

## Decision analysis to select confidence threshold

We needed to find the best trade-off between the effort for manual verification and correction and transcription errors for the *3-digit-sequential* model. We used the confidence scores output from the model for each of the characters in the 3-digit code. Our policy was to send images to manual validation and correction if one of the confidence scores was lower than the threshold. The largest reduction in errors is achieved when manually correcting the first 3%–4%. This corresponds to a confidence threshold of .65 (Figure 7). Further improvement gets more and more costly when we send images that did not pass the threshold value to manual validation and correction, because the total error decreases (Figure 8). The rate at which it does so, however, slows down once we send more than approximately 4%. We therefore set the confidence threshold to 65%. This results in 68,000 images (2.93%) being sent to manual verification and correction, and the estimated accuracy is 97%.

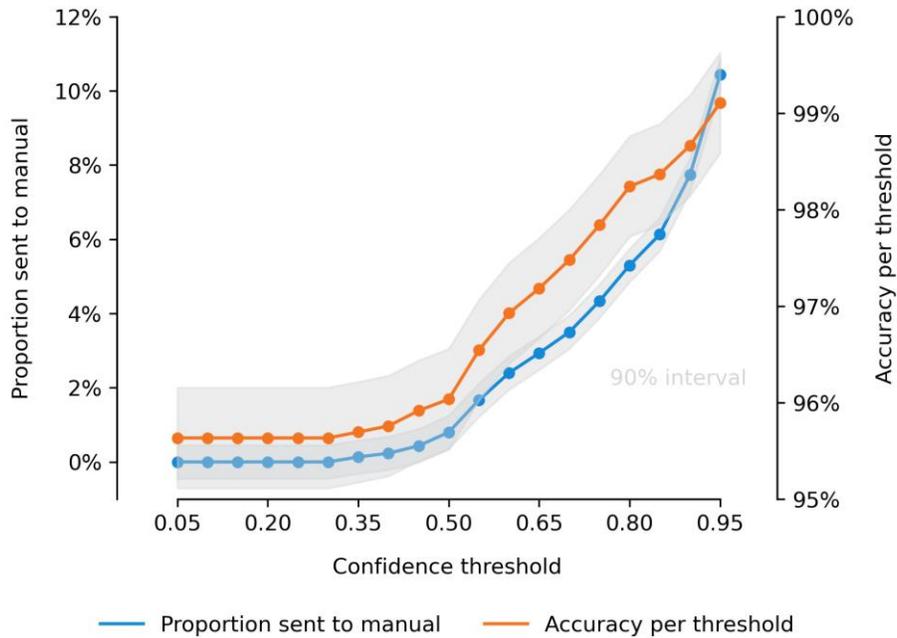

*Figure 7:* *Expected manual verification and correction for different decisions for the 3-digit-sequential model.*

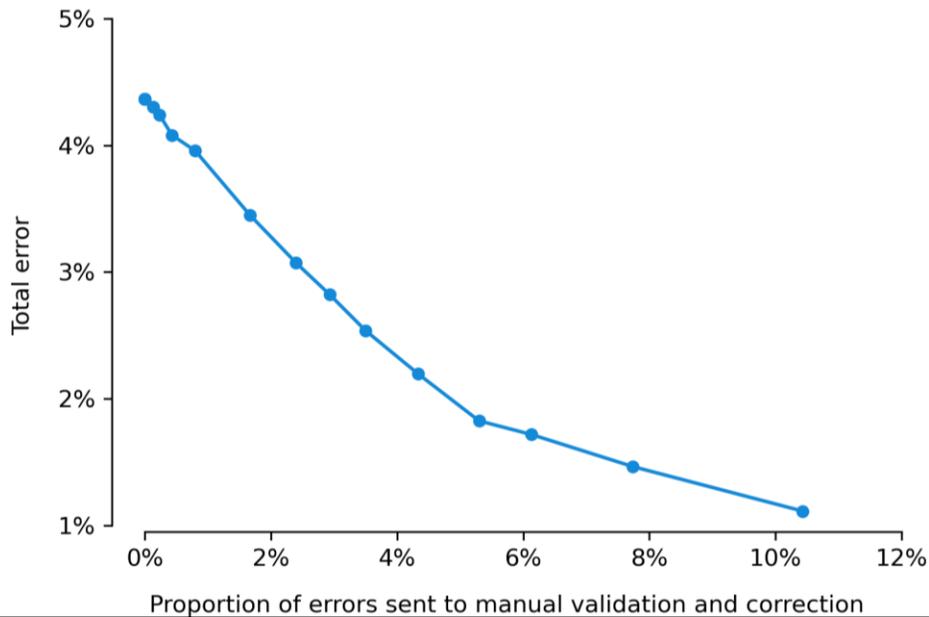

*Figure 8:* *Total error rate per proportion sent to manual validation and correction, 3-digit-sequential model.*

## Evaluation of errors

To evaluate the type of errors produced by the model, we used the 3,000 images in the *verification* dataset. We manually checked the images in the dataset and found that 131 images had been misclassified. The model often misclassified the numbers 1 and 4, or 2 and 8, because these have distinctive features in common, for example the somewhat leaning vertical "stem", or the swooping top loop of the numbers mentioned above. We also found that many errors were caused by noise in the images, such as extra writing added to correct mistakes (Figure 9, left), or the code in the image being unintelligible for both humans and the model (Figure 9, middle). In addition, for some images the human transcriber was wrong, but the model still correctly predicted the code (Figure 9, right).

We also used the *verification* dataset to calculate how many of the images sent to manual verification and correction were correct (Figure 10). Using these results, we estimated that, for the full dataset, 68,000 images (2.93%) would be sent to manual verification and correction using our .65 threshold. Of these, we estimated that 30,138 images (44.3%) would be correctly labelled, and 37,862 images (55.7%) would be incorrectly labelled.

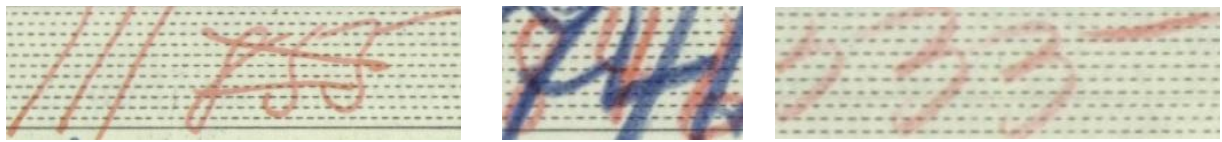

*Figure 9: Examples of misclassified images. Left: an image where the original code has been crossed out and a new one has been entered. Middle: a code that is very hard to read. Right: the human transcriber labelled this image as "533", but the model correctly predicted "555" (the "hat" of the 5s have all been shifted slightly up and to the right, which is not apparent for those unfamiliar with the dataset).*

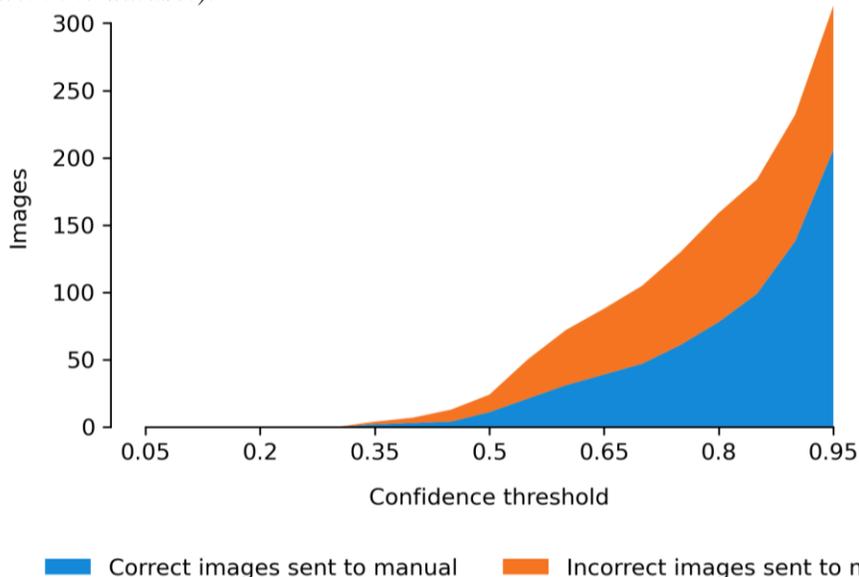

*Figure 10: Verification dataset errors that will be sent to manual verification and correction based on threshold confidences.*

# Discussion

## Lessons learned

We used *Occode* to transcribe the 2.3 million occupation codes in the Norwegian 1950 census. We learned that data processing and data management is very time consuming compared to the development of the machine learning model. We believe our end-to-end pipeline will reduce these efforts. We also learned that pretrained machine learning models did not work well for our data, so we had to label our own data and train our own model. We learned that although a randomly chosen, but frequently used training dataset, had wear and tear on the book pages that introduced artefacts to the images that are not visible for the human eye but that still significantly influenced a deep learning model. The performance achieved for this training dataset did therefore not transfer to the full dataset. We believe this shows the additional challenges of using machine learning in production for large datasets.

A more advanced deep learning architecture, however, worked well for a truly representative training dataset, and the model performance also transferred to the full dataset. We achieved relatively high accuracy (97%) with a manual verification and correction rate (3%) low enough to be useful for production. We have described our analysis for choosing, debugging, tuning and evaluating our models.

## Comparison to related work

Methods to automatically recognize tables in historical documents include machine learning algorithms (Laven, Leishman, & Roweis, 2005) and bottom-up techniques starting from low level features such as table lines to extract the structure. Sibade, Retornaz, Nion, Lerallut, and Kermorvant (2011) deployed a bottom-up technique on a historical French population census (not dated in the publication), where the main goal was to extract first and last names in addition to relation to head of household information for archival indexing. An automation rate of 80% was chosen – that is, rejecting pages with damage to minimize the recognition error rate, ranging from 5.5% to 21%, depending on the page quality. However, their dataset contained a combination of text and digits, and HTR on text is a more difficult problem than on digits. Based on a CNN and RNN trained on a large amount of data, Nion et al. (2013) automatically transcribed nine different columns[2] from the 1930 US Census, with a 70% automation rate and an error rate between 10% and 26%.

Being able to identify the tabular structure of the census questionnaires and extracting the relevant columns were problems we also faced, but instead of utilizing a machine learning approach to solve them, we instead created a script to recognize all horizontal and vertical lines in the image, creating a set of x–y coordinates where these lines intersect. Using these coordinates, we can extract the individual cells from the columns in which we were interested. Using this method, we extract cells from 93% of the questionnaires, although we were not able to automatically detect

---

[2] First and last names, relationship to the head of the family, sex, race, age, marital status, birth place, mother's and father's birth place and immigration year

which cells contained relevant text and which cells did not. We solved this issue by using the CTC loss function when training the model.

Using machine learning to recognize handwritten text in historical documents has been shown to be challenging, because these have overlapping characters across adjacent text lines, show-through and bleed-trough, lack of standard notation, paper texture, ageing, handwriting style, type of pen (dip pen, pencil, ballpoint pen). Models trained on datasets with modern handwriting therefore give poor performance rate when applied to historical documents (Kusetogullari et al., 2019; Romero et al., 2012). For that reason, most handwritten text recognition systems focus on the unique characteristics of a specific historical document with a common layout and style to develop a system that is used to transcribe just these documents (e.g. Thorvaldsen et al., 2015).

In our solution, we started with a training set that had been created by Statistics Norway in such a way as to be representative for the census as a whole. To the human eye, these images did not appear any different from the images from the full census. However, during the project, we began to suspect that there were some differences between these training set images and the remaining images from the census. After creating a binary machine learning model that was trained on the two image sets, it was able to tell them apart. This lends credence to the assertion by Romero et al. (2012) that the condition of historical documents is likely to have deteriorated over time, at least for documents and images that have been in frequent use. Our solution was to create a new training set chosen at random from the full census, where the condition of the images in the training data was therefore much more similar to the full dataset.

Recently, there have been some successful approaches to developing generic systems for historical data transcription. Toledo, Carbonell, Fornés, and Lladós (2019) used known context structures, instead of out-of-vocabulary words so their method does not depend on specific words appearing in training sets. This methodology works well with typical 17$^{th}$ and 18$^{th}$ century parish registers where baptism, funeral and marriage events were registered in books without any tabular structure, but with an expectation that certain information would follow each other, like the first and last name. Similarly, Moysset, Adam, Wolf, and Louradour (2015) described a context-driven model specialized for text line detection, and Chen, Seuret, Hennebert, and Ingold (2017) noted a method for generic page segmentation. While the majority of the literature on line segmentation focuses on detection of the imaginary text lines (i.e. empty lines between lines of actual text), our pre-processing focuses on visible lines, typically found in tabular documents.

We attempted to create a generic solution for digit recognition in images, but we were hampered by the formatting of the cell extraction from the full census pages. Creating and reading the coordinates used for cell extraction is heavily dependent on having them correctly formatted. However, once we come to the part of the pipeline where cell images have been created, the remaining parts of the pipeline (image preprocessing, model creation and training and result analysis) is much more generic and should be easy for others to use by others.

Transkribus (Muehlberger et al., 2019) is a state-of-the-art open source HTR technology that works especially well for unstructured text. When we started this project, Transkribus was not able to extract the structure of the source image. Similar challenges were reported for other projects transcribing historical sources with tabular structure (Lehenmeier, Burghardt, & Mischka, 2020).

However, Transkribus offers a good GUI solution for the manual labelling and exporting of images that is useful for professional transcribers (Vézina, Kermorvant, Bonhomme, & Bournival, 2019).

We achieved best model performance using the CTC loss function that takes an image as input and gives a sequence of the classes found in the image rather than a single classification of an image. A similar approach was taken by Dahl et al. (2021A) in their project to transcribe age, birth and death dates from a collection of Danish death certificates. In their work they modified an attention-based neural network, first proposed by Xu et al. (2015), that takes a single image as input and produces a sequential output and achieved good results.

# Future work

We plan to split the *random sample training* set to single-digit numbers and use these to train one-digit models that we can use for the five single-digit codes in the Norwegian 1950 census (Supplementary Table 1). We will ensure that the occurrence of the digits in the training set is similar to the distribution in the census data. We believe achieving the necessary performance for these columns is easier than for the occupation codes because there are fewer classes. For the 2- and 3-digit columns, we will create and implement a training dataset similar to that used for the *3-digit-sequential* model.

We also plan to evaluate the transferability of our approach to a non-digit column, using the birthplace column from the Norwegian 1950 census. About 2.5 million cells have already been transcribed, including 5,600 different geographical locations.[3] Text is more challenging than digits, because the number of potential classes for the model to give as output increases from 13 to at least 29, and the number of possible class combinations increases drastically. In addition, there will be a greater variance in the quality of the handwriting, because there are longer strings of information, and the individual characters could be more clumped together due to space constraints in the census table. Finally, we assume text columns, such as birthplace, can have an even larger class imbalance than the occupation codes. However, we believe that we can train a model that can be used in production, because we have a very large labelled dataset that will give higher model accuracy and reduce the effect of class imbalance. In addition, we have already demonstrated that our model handles variations in the shape of the characters (digits). However, we expect a higher ratio of the images to be manually verified and corrected. We are therefore addressing the cost of manual verification and correction in our current work. Finally, we can use the approach by Dahl et al. (2021B) to transcribe person names from the HANA database, by using a list of valid place names in Norway at the time, which can then be used to both check the validity of the model results and to find the closest match.

We plan to improve the provenance data management of *Occode*. We plan to use a *big table* (Chang et al., 2008) design to keep track of the data. This is a flexible data structure that can be easily changed by, for example, adding new columns. Currently we are implementing the design using a combination of directory structures in a file system, databases and metadata files. However, we plan to unify the data management in a big table service provided via the cloud in future work. This will make data management easier, improve the performance of data processing and reduce the storage overhead.

---

[3] A 2% sample is manually transcribed, while the remaining material has been semi-automatically transcribed using a GUI for editing similar pictures. The 2.5 million transcribed cells include 65,000 entries registered as "ditto" and 23,000 entries registered with a quotation mark.

We plan to move our data and machine learning pipeline to a commercial cloud platform. A cloud platform enables elastically allocating and paying for the resources needed to transcribe a column. Commercial cloud storage services also provide data lakes (https://azure.microsoft.com/en-us/solutions/data-lake/) and columnar storage systems (like BigTable), which we believe are well suited to our data management needs. They also provide big data processing frameworks such as Spark (Zaharia et al., 2016) and U-SQL (https://docs.microsoft.com/en-us/u-sql/) that we believe are well suited to implementing pipelines such as *Occode*. Finally, we plan to utilize computer vision, natural language processing and other machine learning cloud services to implement transcription for other fields in the 1950 census and other sources.

## Limitations

Our splitter program splits 93% of the questionnaires. We believe most of the remaining questionnaires can be split by iteratively tuning the line detection threshold of the splitter program or by replacing the it with, for example, an object detection and segmentation algorithm such as Detectron2 (https://github.com/facebookresearch/detectron2).

The distribution of occupation codes displayed in Figure 6 (left), shows a long-tailed distribution, meaning that a relatively large proportion of different occupations have few people. Although our automated transcription archives high accuracy, it performs at a lower rate of accuracy for these rare occupation codes. We assume that the manual validation and correction will detect some of the errors, but not all. The two main limitations are (a) in cases where the machine has transcribed the rare code with the code of one of the biggest classes – a not uncommon scenario – and (b) some of the smallest classes are too small to even appear in the training sets. The first issue is more or less impossible to solve, but the latter can be solved with more training data. This raises, however, a question of time and cost efficiency.

We have shown that the distribution of classes in the automatically transcribed dataset is similar, except for the smallest classes. Official statistics from the 1950 census include 'Population by Principal Occupation in the Rural and Town Municipalities and Counties' and 'A Survey of Statistics on Occupation Detailed Figures for the whole Country', and our initial plan was to compare our resulting dataset with the official statistics. However, this proved difficult. The statistics presented in these reports were all constructed based on a combination of individual characteristics and groups of occupational codes that to some extent was not congruent with the codebook used in the statistical preprocessing of the census. Instead, further evaluation of the transcribed results can be done by linking the results with data for the same person in other sources.

# Conclusion

We have described our design and implementation of the *Occode* machine learning pipeline for transcription of historical population censuses. We have used it to automate the transcription of 2.3 million handwritten occupation codes in the Norwegian 1950 population census. Our evaluation shows that it satisfies our three requirements: (a) the accuracy is 97%, (b) only 3% of the codes requires manual verification, and (c) the distribution of codes is correct, except for the smallest classes. The transcribed results will be implemented in the Norwegian Historical Population register.

Our main reason for using automated transcription, including model tuning, is to improve cost-efficiency. Although the cost of creating this end-to-end pipeline was initially high, we have achieved an accuracy that allows us to use the generic pipeline to also transcribe other digit columns from both the 1950 census and other censuses with a small amount of work. We therefore assume that our development cost is negligible when compared to the cost of manual transcription for these fields. We believe our approach and the lessons learned are useful for other transcription projects that plan to use machine learning in production. The pipeline source code is open source and available at: https://github.com/uit-hdl/rhd-codes. We have also published our labelled training dataset for the 3-digit occupation codes: https://doi.org/10.18710/OYIH83 (*2% dataset*) https://doi.org/10.18710/7JWAZX (*random dataset*). We have uploaded the source code at time of publication, trained model, as well as the confidence scores produced by model: https://doi.org/10.6084/m9.figshare.14573325. The automatically transcribed codes will be included into the Norwegian Historical Population Register.

# Acknowledgements

Thanks to the Målselv transcription team at the Norwegian Historical Data Centre. Also, thanks to Gunnar Thorvaldsen and Tim Alexander Teige, who initiated the project that led to *Occode*. We acknowledge funding provided by UiT the Arctic University of Norway through the interdisciplinary strategic project High North Population Studies and funding provided by the Research Council of Norway (225950/F50).

# Supplementary materials

***Supplementary Table 1****: Other columns with numeric codes in the Norwegian 1950 census.*

| Column | Number of digits | Valid range | Distribution source |
| --- | --- | --- | --- |
| Position in household | 1 | 1-3 | Central statistics |
| Resident | 1 | 1-5 | Central statistics |
| Marriage status | 1 | 1-7 | Central statistics |
| Employment | 1 | 1-9 | Central statistics |
| Business | 3 | 500 codes | Central statistics |
| Lower education | 1 | 1-9 | Central statistics |
| Higher education | 3 | 163 codes | Central statistics |
| Nationality | 2 | 01-50 | Central statistics |

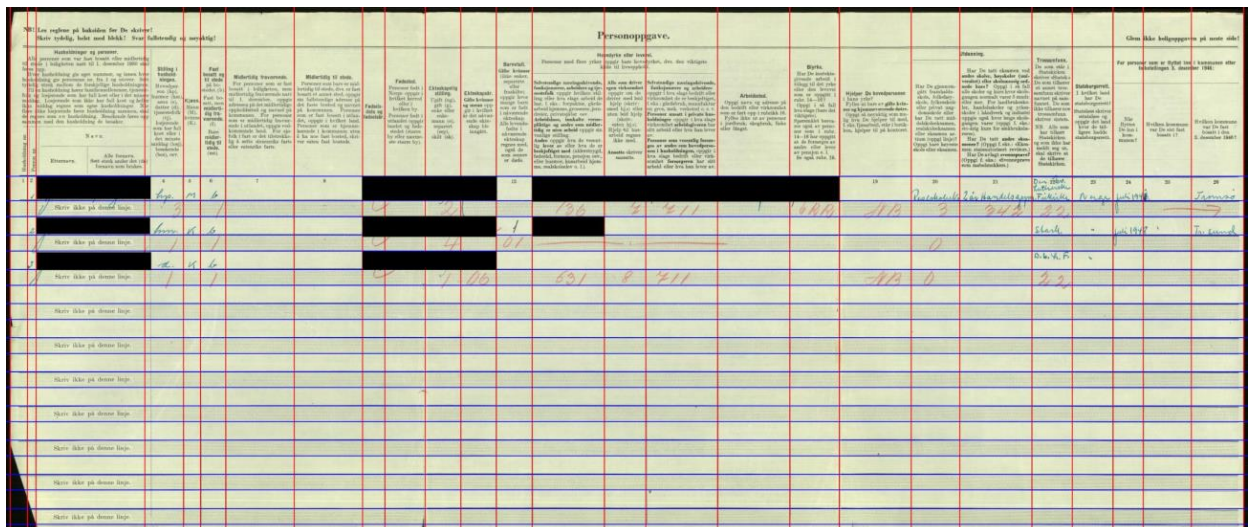

***Supplementary figure 1***: *"Personoppgave" (questionnaires) from the 1950 Norwegian population census split into individual cells (shown by red and blue lines). We have censored sensitive information that we do not have permission to publish (black boxes).*